\documentclass[conference]{IEEEtran}

\usepackage{cite}
\usepackage{amsmath,amssymb,amsfonts}
\usepackage{algorithmic}
\usepackage{graphicx}
\usepackage{textcomp}
\usepackage{xcolor}
\usepackage{booktabs}
\usepackage{multirow}
\usepackage{subcaption}
\usepackage[colorlinks=true, citecolor=black, linkcolor=black, urlcolor=black]{hyperref}
\def\adam/{Adam}
\def\bert/{bert-base-cased}
\def\fbank/{mel-filterbank}
\def\llama/{Llama}
\def\llamathree/{Llama-3.1-8B-Instruct}
\def\qwen/{Qwen2}
\def\qwentwo/{Qwen2-7B-Instruct}
\def\effnet/{EfficientNet}
\def\whisperl/{whisper-large-v2}
\def\demenba/{Demenba}
\newcommand{\liming}[1]{}

\def\BibTeX{{\rm B\kern-.05em{\sc i\kern-.025em b}\kern-.08em
    T\kern-.1667em\lower.7ex\hbox{E}\kern-.125emX}}
\begin{document}

\title{Recognizing Dementia from Neuropsychological Tests with State Space Models
}
\author{Liming Wang$^1$, Saurabhchand Bhati$^1$, Cody Karjadi$^2$, Rhoda Au$^2$, James Glass$^1$\\
$^1$\textit{MIT CSAIL}, $^2$\textit{Boston University}\\
\texttt{limingw@csail.mit.edu}
}

\maketitle

\begin{abstract}
\liming{Make sure it is less than 1000 characters}
Early detection of dementia is critical for timely medical intervention and improved patient outcomes. Neuropsychological tests are widely used for cognitive assessment but have traditionally relied on manual scoring. Automatic dementia classification (ADC) systems aim to infer cognitive decline directly from speech recordings of such tests. We propose \demenba/, a novel ADC framework based on state space models, which scale linearly in memory and computation with sequence length. Trained on over 1,000 hours of cognitive assessments administered to Framingham Heart Study participants, some of whom were diagnosed with dementia through adjudicated review, our method outperforms prior approaches in fine-grained dementia classification by 21\%, while using fewer parameters. We further analyze its scaling behavior and demonstrate that our model gains additional improvement when fused with large language models, paving the way for more transparent and scalable dementia assessment tools.\footnote{Code: https://anonymous.4open.science/r/Demenba-0861}

\end{abstract}


\section{Introduction}

\vspace{1em}
Dementia, including Alzheimer's disease, is a progressive neurodegenerative condition that severely impairs memory, reasoning, communication, and other cognitive functions. 
As the global population ages, early diagnosis of dementia has become increasingly important for initiating timely treatment, slowing disease progression, and improving patients' quality of life and life expectancy~\cite{szekely2004nsaids,chuang2016midlife}.

A common clinical approach for assessing cognitive impairment is the neuropsychological test, such as the widely used Mini-Mental State Examination (MMSE)~\cite{kurlowicz1999mmse}. In these tests, trained clinicians guide patients through a series of structured tasks evaluating memory, attention, perception, verbal fluency and reasoning. 
While effective, this process is labor-intensive, prone to subjective biases and often inconsistent, especially in distinguishing subtle or early-stage cognitive decline~\cite{Luz2021_alzheimer_adresso}. Furthermore, manual scoring may miss nuanced acoustic or linguistic indicators that signal cognitive deterioration.
To alleviate these issues, automatic dementia classification (ADC) systems have been developed to infer cognitive status directly from neuropsychological tests~\cite{alhanai2017spoken,Luz2021_alzheimer_adresso,dawalatabad-etal-2022-detecting}. By analyzing speech recordings, these systems aim to detect subtle linguistic and paralinguistic cues (e.g., hesitations, disfluencies, semantic anomalies) indicative of dementia. In addition to reducing clinician burden, such systems offer the potential to standardize assessment and mitigate biases, including those introduced by leading or suggestive questions~\cite{al-hanai-etal-2018-role,dawalatabad-etal-2022-detecting,perez-toro2021interviewer}.

However, existing ADC systems are fundamentally limited in their ability to process long neuropsychological test recordings --- typically around one hour in duration. Most current models, particularly those based on transformer architectures~\cite{Vaswani2017}, struggle to handle more than 30 seconds of audio at a time due to quadratic growth in memory and computation with input length~\cite{gu2021efficiently}. This constraint often forces segment-level inference using forced alignment or manual heuristics~\cite{dawalatabad-etal-2022-detecting,Balagopalan2021-alzheimer-w2v2,Li2023-alzheimer-whisper}, leading to context fragmentation and a drop in fine-grained classification performance~\cite{bhati2024dass}.

Alternatively, transcription-based pipelines using an automatic speech recognizer (ASR) suffer from loss of acoustic information as well as error propagation, especially in  noisy, spontaneous and multi-speaker conversational settings. 

To address these challenges, we propose leveraging state-space models (SSMs)~\cite{gu2021efficiently,gu2022efficiently,gu2023mamba}, a family of architectures designed for efficient long-sequence modeling. Unlike transformers, SSMs scale linearly in both space and time, making them ideal for modeling full-length interviews without segmentation. Furthermore, the dementia information in the neuropsychological tests can be subtle and sporadic, with many conversational turns offering little diagnostic value~\cite{al-hanai-etal-2018-role}. SSMs' natural capacity for temporal compression allows them to distill salient patterns with minimal information loss, making them particularly well-suited for the ADC task. In this paper, we make three main contributions: 
\begin{enumerate}
    \item We present \emph{\demenba/}, a memory- and compute-efficient architecture trained on over 1,000 hours of neuropsychological tests with balanced representation across dementia stages;
    \item Our method achieves superior ADC accuracy compared to the state-of-the-art model~\cite{dawalatabad-etal-2022-detecting} by up to 21\% relative AUC, particularly in fine-grained classification settings (e.g., mild cognitive impairment vs. dementia). It also requires significantly fewer trainable parameters. The performance of our method further improves after fusing with text-based large language models (LLMs);
    \item We conduct extensive ablation studies to assess the contribution of different speech segments. Our scaling experiments further suggest that \demenba/ maintains robust performance as both data and model size change.  
\end{enumerate}

\section{Related work}

\liming{CITE MORE and figure out the citation style issue}
Early work on ADC tasks such as Alzheimer detection (AD) used classical machine learning algorithms with hand-crafted speech and linguistic features~\cite{alhanai2017spoken,Luz2020_alzheimer_adress}. More recent systems leverage deep learning architectures such as convolutional~\cite{dawalatabad-etal-2022-detecting} and recurrent~\cite{Rohanian2021-alzheimer-speech,xue2021dementia} neural networks and neural embeddings from pretrained speech representation models such as wav2vec 2.0~\cite{Baevski2020-wav2vec2,Balagopalan2021-alzheimer-w2v2} and Whisper~\cite{Radford2023-whisper,Li2023-alzheimer-whisper} as well as text language models~\cite{Haulcy2021-alzheimer-speech,Li2023-alzheimer-whisper}. Despite progress in algorithmic design, existing work still focuses on sentence-level speech segments and small datasets such as Alzheimer’s Dementia Recognition through Spontaneous Speech (ADReSS)~\cite{Luz2020_alzheimer_adress,Luz2021_alzheimer_adresso} and Fragmingham Heart Study (FHS) 92-hour subset~\cite{alhanai2017spoken} with less than 30 hours of dementia speech combined due to privacy concerns. Among the works, \cite{al-hanai-etal-2018-role,perez-toro2021interviewer,dawalatabad-etal-2022-detecting} included examiner speech in their study and surprisingly found that AD is possible with examiner speech only, indicating examiner bias.

State-space models (SSMs) are proposed as more memory- and compute-efficient alternatives to transformers~\cite{Vaswani2017}, especially when processing long sequences~\cite{gu2021efficiently,gu2022efficiently,gu2023mamba}. While earlier SSMs are auto-regressive with linear recurrent layers~\cite{gu2021efficiently}, later variants improve their expressivity by incorporating data-dependent selective scan block~\cite{gu2023mamba,liu2024vmamba}. To further increase their modeling capacity, particularly of two-dimensional data such as vision, bidirectional connections from left-to-right~\cite{zhu2024vision} and multidirectional connections from left-to-right and top-to-bottom~\cite{liu2024vmamba} have been proposed. 
SSMs have been successfully applied to the audio signal in tasks such as audio event classification~\cite{erol2024audio,lin2024audio,bhati2024dass} and self-supervised audio representation learning~\cite{shams2024ssamba}. Knowledge distillation~\cite{bhati2024dass} from transformer teachers can help boost SSMs' performance and even outperform the transformer teachers. SSMs have also been used for efficient automatic speech recognition~\cite{zhang2025mamba, jiang2025speech}, separation, and  synthesis~\cite{jiang2025speech}.

\section{Method}
\begin{figure}
    \centering
    \includegraphics[width=0.4\textwidth]{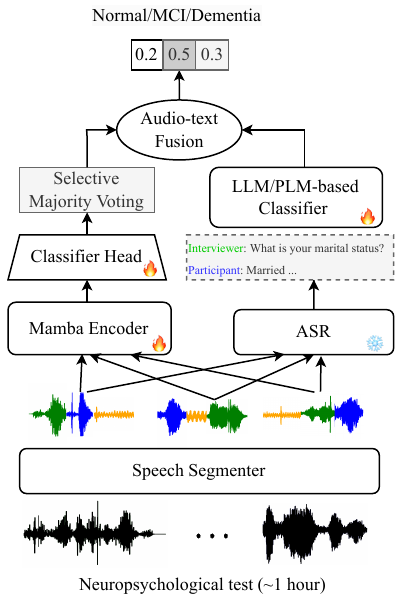}
    \caption{\textbf{Overall architecture of the proposed ADC model}. The model consists of a frozen speech segmenter that divides the hour-long recording into shorter segments, a trainable SSM-based audio classifier and a trainable text-based text classifier. The predictions from the two classifiers are then combined via late fusion.}
    \label{fig:arch}
\end{figure}
The overall architecture of our SSM-based ADC system is illustrated in Fig.~\ref{fig:arch}. The complete system consists of four main components: a speech segmenter, an automatic speech recognizer (ASR), an audio-based dementia classifier, and a text-based dementia classifier. First, the speech segmenter divides each hour-long recording into shorter chunks using either a simple voice activity detector (VAD) or a more fine-grained speaker diarization model (SD). To preserve longer context while respecting a fixed memory budget, adjacent chunks are merged into segments up to a predefined maximum length (e.g., six minutes). Each merged segment is represented by its \fbank/, which serve as input to the audio classifier. To handle long segments, we employ Mamba~\cite{gu2023mamba} SSM backbone, retaining the exactly the same model parameters as in the setup of \cite{bhati2024dass}. Concretely, the backbone consists of four groups of Mamba layers, each followed by a downsampling factor of two; the final output is mean-pooled to produce class logits. Training is performed with a cross entropy loss. During inference, we observe that treating every segment equally (i.e., standard majority voting) allows noisy or uninformative segments to dilute the overall decision. To mitigate this, we adopt a \emph{selective} majority voting scheme. After computing segment-level probabilities $p^1_{\text{audio}},\cdots,p^N_{\text{audio}}$ for all $N$ segments, we choose the top $k$ segments by highest peak probability (where $k$ is tuned on a held-out validation set). A soft majority vote over these $k$ segments using the estimated class probabilities produce the recording-level score $p_{\text{audio}}$. In preliminary experiments, this selective voting consistently outperformed na\"{i}ve majority voting  over all segments by filtering out segments that are neither indicative of dementia nor contain reliable audio cues.   

In parallel, the SD output is fed to the ASR to obtain speaker-tagged transcripts: for each utterance, we prepend the speaker label (e.g., ``Participant'' or ``Interviewer'') to the transcribed text, yielding a complete transcript for the recording. The text classifier then consumes this concatenated transcript to produce a text-based probability $p_{\text{text}}$. We explore both zero-shot classification with large language models (LLMs) and finetuning approaches using either LLMs or pretrained LMs (PLMs) such as BERT~\cite{devlin2019bert}. For zero-shot LLM inference, we construct a structural prompt that includes:
\begin{itemize}
    \item \emph{Task description}: ``You are a helpful assistant that classifies if a participant in an interview has dementia'';
    \item \emph{Interview transcript}: each line formatted as \texttt{[speaker role]}:\texttt{[transcript]}, where \texttt{[speaker role]} is ``Participant'' or ``Interviewer''; 
    \item \emph{Label set}: a list valid labels for the task, e.g., ``normal, dementia'' for 2-class classification.
    \item \emph{Instruction}: simply ``Answer:''.
\end{itemize}
We then extract the logit corresponding to the first token of the model's response and map it to a probability score over the two labels. Because most LLMs have a strict context-length limit, we either (a) feed the entire transcript into the model at once -- if it fits within the context window -- or (b) apply the same selective majority voting scheme as in the audio branch. For fine-tuning experiments (both with LLMs and PLMs), we only use selective majority voting to reduce GPU memory usage: we fine-tune on individual segments, select the top $k$ segment scores at inference time, and then combine them. Lastly, we fuse the audio and text scores via
\begin{align}
    p_{\text{audio+text}}=(1-\lambda)p_{\text{audio}}+\lambda p_{\text{text}},
\end{align}
where the \emph{fusion weight} $\lambda\in[0,1].$ This late fusion allows the system to leverage complementary evidence from both modalities while controlling for over-reliance on either branch.

\section{Experiment}

We evaluate the models on tasks including dementia detection and the more fine-grained 3-class dementia classification on the Framingham Heart Study (FHS) dataset~\cite{al-hanai-etal-2018-role}. The details are described in the next sections.
\begin{table}[ht]
    \centering
    \caption{Statistics of the FHS dataset. N/M/D stands for the number of recordings for normal/MCI/dementia participants.}
    \begin{tabular}{lccccc}
    \toprule
        split &  \# participants & \# interviewers & N/M/D & \# hours \\
    \midrule
       train (2-class)  & 378 & 61 & 400/234/173 & 943\\
       train (3-class)  & 586 & 74 & 407/399/399 & 1447 \\
       test (2-class) & 10 & 11 & 10/5/5 & 25 \\
       test (3-class) & 18 & 11 & 10/10/10 & 36 \\
       eval & 92 & 20 & 68/10/14 & 77 \\ 
    \bottomrule
    \end{tabular}
    \label{tab:data}
\end{table}
\subsection{Dataset and training settings}\label{sec:data}
We conduct all experiments on the Framingham Heart Study (FHS) dataset~\cite{al-hanai-etal-2018-role}, which consists of approximately 11,000 hour-long neuropsychological test recordings, 2,058 of which have been reviewed for dementia. After adjudication, each recording is labeled as one of three classes --- normal, mild cognitive impairment (MCI) and dementia --- of which 936 are normal. To balance the dataset, we randomly select roughly 400 recordings per class, yielding 1,200 recordings in total. We consider two classification settings:
\begin{itemize}
    \item \emph{3-class} classification using the original labels.
    \item \emph{2-class} classification by merging MCI and dementia into a single ``impaired'' class and randomly sampling 400 recordings from the combined 800 recordings.   
\end{itemize}
In both settings, the test set comprises 10 recordings per class, sampled randomly. We ensure that participants do not overlap between the training, test splits. Additional dataset statistics are provided in Table~\ref{tab:data}. To segment each recording, we employ two approaches: 
\begin{itemize}
    \item A coarse-grained segmentation using a VAD\footnote{\href{https://github.com/wiseman/py-webrtcvad.git}{https://github.com/wiseman/py-webrtcvad.git}} to split the recordings into silence and speech segments.
    \item A fine-grained segmentation using an SD system from the pyannote toolkit~\cite{Plaquet23,Bredin23}.
\end{itemize}
 In the latter approach, we feed the entire speech waveform into the diarizer and label the first detected speaker with a segment duration more than 20ms as the interviewer, and treat all the subsequent speakers as the participant. For both methods, we merge consecutive segments --- up to a maximum duration of 360 seconds --- to form the model input. For text-based classifiers, we apply Whisper-Large v2 ASR~\cite{Radford2023-whisper} to each diarized speech segment. To compare with prior work, we also evaluate our approach on a disjoint 92-recording subset of FHS~\cite{alhanai2017spoken,dawalatabad-etal-2022-detecting} that contains manual transcripts and speaker diarization labels. On this subset, the ASR yields a character error rate of approximately 60\%, reflecting prevalent disfluencies and noisy recording conditions. We also inspected the frequency of dementia-related keywords in our training and test sets, finding 83 occurrences of ``Alzheimer'' and 72 occurrences of ``dementia'' in the training set, and only one occurrence of ``Alzheimer'' and zero occurences ``dementia'' in the test set. By examining the contexts of these mentions, we confirmed that none directly reveal participants' cognitive status, consistent with standard neuropsychological testing protocols that prohibit interviewers from disclosing such information~\cite{al-hanai-etal-2018-role}.
 
We follow prior work~\cite{dawalatabad-etal-2022-detecting} and extract 128-bin \fbank/s with a 10ms frame shift, a 25ms frame length, and a Hanning window.  
Our audio classifier is based on VMamba~\cite{liu2024vmamba}\liming{Confirm}, configured exactly as in DASS~\cite{bhati2024dass} and initialized with ImageNet~\cite{deng2009imagenet}-pretrained weights, which outperform an AudioSet~\cite{gemmeke2017audio}-pretrained alternative in preliminary experiments. We refer to the larger model as \demenba/-medium, corresponding to DASS-medium~\cite{bhati2024dass}, and to the smaller model as \demenba/-small, sharing its architecture with DASS-small~\cite{bhati2024dass}. For comparison, we also implement the previous state-of-the-art \effnet/-based model~\cite{dawalatabad-etal-2022-detecting}, using the b6 variant~\cite{tan2019efficientnet}, which is the largest variant that fits our GPU memory constraints when processing 360-second segments. All models share the same input features, segment durations, and optimization hyperparameters such as batch size and training epochs. For the text classifier, we experiment with two families of text models: PLMs such as BERT~\cite{devlin2019bert}, with a three-layer multilayer perceptron (MLP) of 768 hidden units appended on top of BERT's final embedding layer, and LLMs, including \llama/~\cite{touvron2023llama}, \qwen/~\cite{yang2024_qwen2} and phi-4~\cite{phi4mini2024technical}. For each LLM, we apply a low-rank adaptor (LoRA)~\cite{hu2021lora} of rank 8 to every layer for finetuning.
In the 2-class setting, we compare binary cross entropy (BCE) and (multinomial) cross entropy (CE) losses, finding that CE works best for \demenba/, whereas BCE performs better for \effnet/. In the 3-class setting, we employ a weighted CE loss with weights $(1,3,3)$ (normal, MCI, dementia), which yields the highest validation performance. 
We train each \demenba/ model for 40 epochs with Adam \cite{kingma2015adam}, using an initial learning rate of $10^{-5}$, $\beta_1=0.95$, $\beta_2=0.999$, weight decay $=5\times10^{-7}$, and a batch size of 1 (to fit the longest segments). We warm up for 1,000 steps, then apply an exponential decay of 0.5 per epoch starting at epoch 10. All training is done on a single NVIDIA A6000 GPU (48 GB).
We measure performance using the Area Under the Receiver Operating Characteristic Curve (AUC), which enables comparison across all detection thresholds. Results from the best
models on the test set are reported, including those on the eval set.
\begin{table}[t]
    \centering
    \caption{\textbf{Dementia classification results for audio-only models}. We compare the proposed methods with the state-of-the-art method based on EfficientNet~\cite{dawalatabad-etal-2022-detecting}. only AUCs for the better of BCE and CE losses are shown. All models are trained on the 400 hour/class datasets with \fbank/ features of 360-second segments as inputs. $(3\rightarrow 2)$ means results from a 3-class classifier by merging MCI+dementia probabilities.}
    \resizebox{0.48\textwidth}{!}{
    \begin{tabular}{lcccc}
    \toprule
          &  \begin{tabular}{@{}c@{}}
               \textbf{\# Trainable}\\
               \textbf{Param.}
          \end{tabular}& \begin{tabular}{@{}c@{}}
               \textbf{Segment}\\
               \textbf{Boundary}
          \end{tabular} & \textbf{Loss} & \textbf{AUC ($\uparrow$)} \\
    \midrule
    \midrule
    \multicolumn{5}{c}{\emph{2-class Classification}} \\
    \midrule
    \midrule
     \multirow{2}{*}{\effnet/ b6} & \multirow{2}{*}{40m} & VAD & BCE & 0.76 \\
      &  & SD & BCE & 0.82 \\
     \midrule
     \multirow{2}{*}{\demenba/-small} & \multirow{2}{*}{29m} & VAD & CE & 0.77\\
     & & SD & CE & \underline{0.85}\\
     \demenba/-small $(3\rightarrow 2)$\liming{Discuss} & 29m & SD & CE & 0.82 \\
     \multirow{2}{*}{\demenba/-medium} & \multirow{2}{*}{48m} & VAD & CE & 0.82 \\
      & & SD & CE & \textbf{0.87} \\
     \midrule
     \midrule
     \multicolumn{5}{c}{\emph{3-class Classification}} \\
     \midrule
     \midrule
     \multirow{2}{*}{\effnet/ b6} & \multirow{2}{*}{40m} & VAD & CE & 0.62 \\
     & & SD & CE & 0.69 \\
     \midrule
     \multirow{2}{*}{\demenba/-small} & \multirow{2}{*}{29m} & VAD & CE & 0.68 \\
     & & SD & CE & \underline{0.81}\\
     \multirow{2}{*}{\demenba/-medium} & \multirow{2}{*}{48m} & VAD & CE & 0.75 \\
     & & SD & CE & \textbf{0.83} \\
    \bottomrule
    \end{tabular}}
    \label{tab:result_audio}
\end{table}
\subsection{Overall results}
Table~\ref{tab:result_audio} summarizes our audio-only ADC performance. In the 2-class setting, \demenba/-small achieves an AUC comparable to or better than the \effnet/ baseline with significantly fewer parameters. \demenba/-medium surpasses \effnet/ by 6\% and 5\% absolute AUC using VAD and SD boundaries respectively, despite incurring only a modest parameter increase. For 3-class classification, \demenba/-small and \demenba/-medium improve over \effnet/ by 6\% and 13\% AUCs using VAD boundaries and 12\% and 14\% AUCs using SD boundaries  respectively. This gap grows as the task becomes more fine-grained, underscoring the benefit of SSMs when discriminating subtle differences between MCI and dementia. Among the \demenba/ classifier, medium performs better than small by 10\% absolute AUC for 2-class and 7\% absolute for 3-class, demonstrating the scalability of our method. We also compare two segmentation strategies: a coarse-grained VAD versus a fine-grained SD. Across all models and both classification settings, SD-based boundaries outperform VAD by as much as 5\% absolute AUC. Note that we control irrelevant variables such as the segment length to ensure the difference is not due to change in segment length. Instead, SD preserves the within-speaker conversational context (turn-taking, prosodic continuity), which appears crucial for ADC. Lastly, 
we observed that direct 2-class finetuning outperforms training a 3-class model and merging MCI+dementia outputs for 2-class evaluation by 3 AUC points. 
\begin{table}[t]
    \centering
    \caption{\textbf{Dementia classification results for text-only and audio+text models.} X+BERT models use the fully finetuned \bert/ (with an MLP head) as the text classifier. X+\llama/ models use the finetuned \llamathree/ as the text-based classifier and X+\qwen/ models use the finetuned \qwentwo/ as the text classifier. All audio models use segment boundaries from an SD. ``Segment length=Full'' means the whole the recording is fed into the model in one forward pass.}
    \resizebox{0.48\textwidth}{!}{
    \begin{tabular}{lcccc}
    \toprule
         & \begin{tabular}{@{}c@{}}
            \textbf{\# Trainable}\\\textbf{Param.}
        \end{tabular} & \begin{tabular}{@{}c@{}}
            \textbf{Finetuning}\\
             \textbf{Method}
        \end{tabular} & \begin{tabular}{@{}c@{}}
          \textbf{Segment}\\
          \textbf{Length (s)}
          \end{tabular} & \textbf{AUC ($\uparrow$)}\\
    \midrule
    \midrule
    \multicolumn{5}{c}{\emph{2-class Classification}} \\
    \midrule
    \midrule
    \multirow{1}{*}{\bert/} & 109m & Full & 180 & \underline{0.91} \\
    \midrule
    \multirow{3}{*}{\llamathree/} & 0 & No & Full & 0.73 \\
     & 0 & No & 360 & 0.53 \\
     & 21m & LoRA & 360 & 0.83 \\
    \midrule
    \multirow{3}{*}{\qwentwo/} & 0 & No & Full & 0.60 \\
     & 0 & No & 360 & 0.80 \\
     & 20m & LoRA & 360 & 0.85 \\
    \midrule
    \multirow{2}{*}{phi-4} & 0 & No & Full & 0.64 \\
     & 0 & No & 360 & 0.65 \\
    \midrule
    \demenba/-medium+BERT & 157m & Full & 180 & \textbf{0.95} \\
    \demenba/-medium+\llama/ & \multirow{1}{*}{69m} & LoRA & 360 & 0.90\\
    \demenba/-medium+\qwen/ & \multirow{1}{*}{68m} & LoRA & 360 & 0.87\\
    \demenba/-small+\llama/ & \multirow{1}{*}{68m} & LoRA & 360 & 0.87\\
    \demenba/-small+\qwen/ & \multirow{1}{*}{68m} & LoRA & 360 & 0.85\\
    \bottomrule
    \end{tabular}
    }
    \label{tab:result_audio_text}
\end{table}

\begin{table}[t]
    \centering
    \caption{\textbf{Dementia classification results on the eval set using the best checkpoints from the test set.} X+BERT models use the fully finetuned \bert/ (with an MLP head) as the text-based classifier. ``m'' stands for million and ``b'' stands for billion. All models use  ground truth speaker segment boundaries. $\sim$ denotes average length and without $\sim$ denotes maximal length.}
    \resizebox{0.48\textwidth}{!}{
    \begin{tabular}{lcccc}
    \toprule
         & \begin{tabular}{@{}c@{}}
            \textbf{\# Trainable}\\\textbf{Param.}
        \end{tabular} & \begin{tabular}{@{}c@{}}
            \textbf{Input}\\
             \textbf{Type}
        \end{tabular} & \begin{tabular}{@{}c@{}}
          \textbf{Segment}\\
          \textbf{Length (s)}
          \end{tabular} & \textbf{AUC ($\uparrow$)}\\
    \midrule
    \midrule
    \multicolumn{5}{c}{\emph{2-class Classification}} \\
    \midrule
    \midrule
    \multirow{1}{*}{\effnet/ ~\cite{dawalatabad-etal-2022-detecting}} & - & Audio & $\sim$8 & 0.78 \\
    \multirow{1}{*}{\effnet/+LM~\cite{dawalatabad-etal-2022-detecting}} & - & Audio+Text & $\sim$8 & 0.83 \\
    \midrule
    \multirow{2}{*}{\bert/} & \multirow{2}{*}{109m} & Text & 180 & 0.88 \\
   & & ASR text & 180 & 0.75 \\
    \multirow{1}{*}{\demenba/-medium} & 48m & Audio & 360 & \textbf{0.81} \\
    \multirow{1}{*}{\demenba/-small} & 29m & Audio & 360 & 0.71 \\
    \multirow{1}{*}{\demenba/-medium+BERT} & 157m & Audio+Text & 360 & \textbf{0.92} \\
    \bottomrule
    \end{tabular}
    }
    \label{tab:result_gold92}
\end{table}
Table~\ref{tab:result_audio_text} presents performance for text-only and audio+text hybrids. In the 2-class setting, \bert/ achieves an AUC of 0.91 --- outperforming the best finetuned LLM (\qwentwo/) by $6\%$ absolute. Although recent LLMs show strong general language understanding, most of them perform poorly zero-shot for ADC, likely because of the high error rate of the ASR transcripts. One exception is \qwentwo/, which achieves an AUC of 0.80 in the zero-shot setting, showing that the LLM has some level of understanding about dementia. 
Performing fine-tuning on LLMs further narrows the gap but does not surpass BERT, which suggests that, under noisy transcript conditions, a mid-sized PLM fine-tuned end-to-end remains the best  text approach for ADC. Importantly, combining modalities yields further gains: the hybrid of BERT+\demenba/-medium achieves an AUC of 0.95, the single best result across all systems. This synergy indicates that audio-based and text-based classifiers extract complementary features: audio models capture prosodic cues such as hesitation and intonation, whereas text models capture lexical/linguistic patterns like filler words and semantic incoherence. 

Table~\ref{tab:result_gold92} reports evaluation on the 92-recording dataset with manual transcripts and manual diarization. We benchmark our audio-only \demenba/-medium and audio+text \demenba/-medium+BERT against \effnet/ and \effnet/+LM. On this subset, \demenba/-medium alone outperforms \effnet/ by 3\% absolute AUC, and \demenba/-medium+BERT outperforms \effnet/+LM by 9\% absolute AUC, demonstrating the advantage of leveraging longer context. We also found that BERT-based text classifier performs better using real text than using ASR text by 13\% absolute AUC, despite being trained on ASR text. One possible reason is that the eval set contains shorter and noisier recordings, resulting in domain mismatch between training and evaluation. The larger AUC gap between \demenba/-small and -medium also suggests \demenba/ to be more generalizable and robust to change in recording conditions.

\begin{figure}[t]
    \centering
    \begin{subfigure}{0.46\textwidth}
        \centering
        \includegraphics[width=0.95\textwidth]{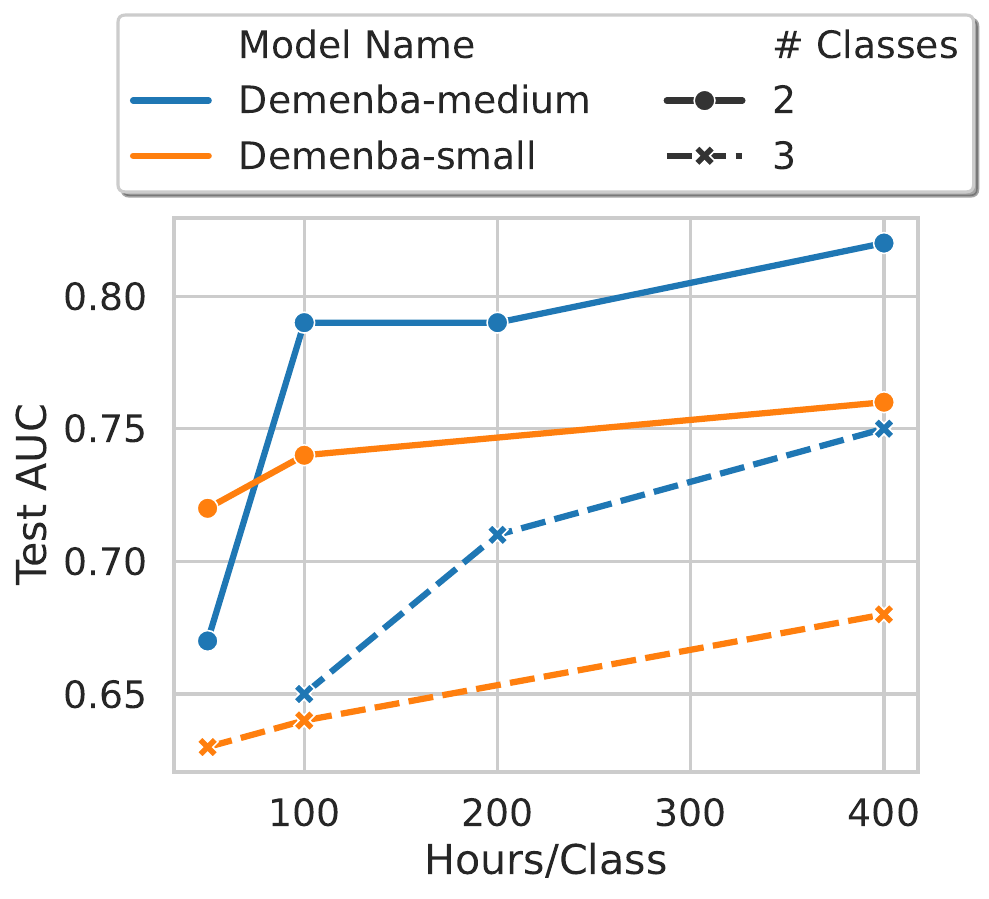}
        \caption{Classification AUC vs. training sample size}
        \label{fig:eff_of_sample_size}
    \end{subfigure}
    \begin{subfigure}{0.46\textwidth}
        \centering
        \includegraphics[width=0.95\textwidth]{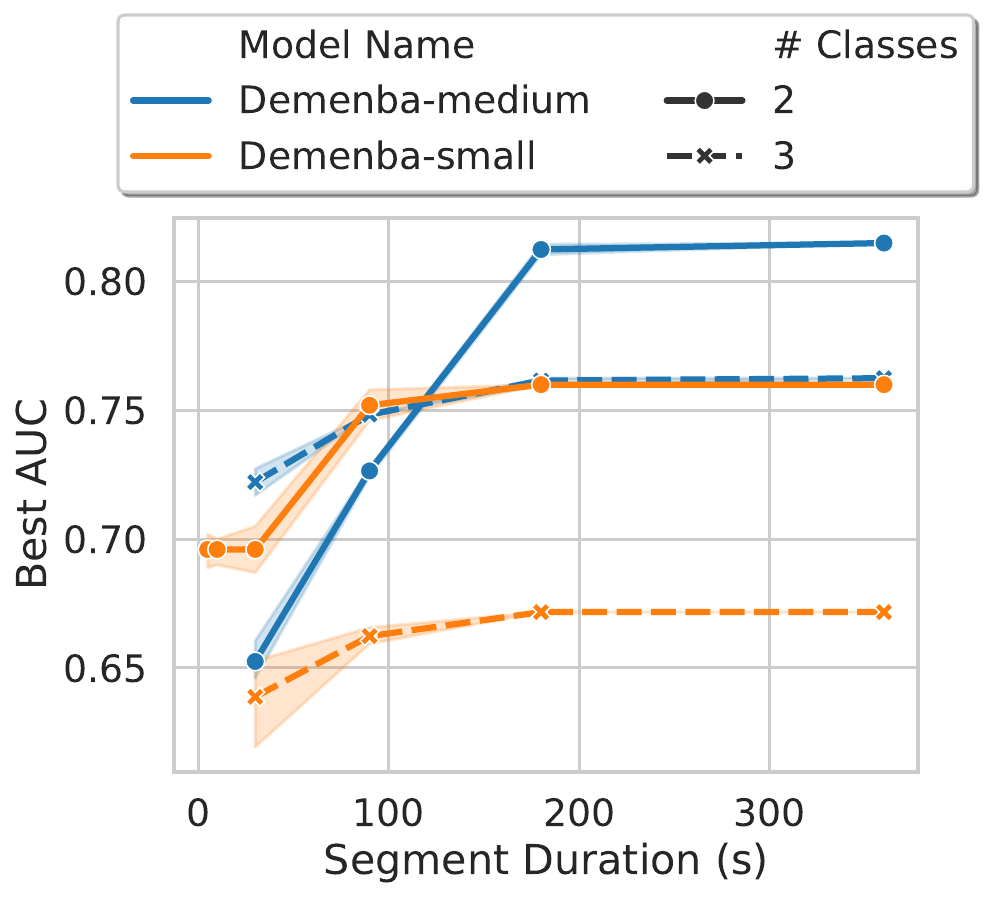}
        \caption{Classification AUC vs. input segment length}
        \label{fig:eff_of_segment_duration}
    \end{subfigure}
    \caption{\textbf{Scaling behavior of \demenba/ with sample size and segment duration.} (a) Scaling behavior with training sample size. All models take 360-second segments as inputs; (b) Scaling behavior with test segment duration. All models are trained on the 400-hour/class datasets. Best AUC is the highest AUC for all segment durations up to the current value.\liming{Confidence interval}}
    \label{fig:eff_of_scaling}
\end{figure}
\subsection{Scaling behavior: sample size and sequence length}
Fig.~\ref{fig:eff_of_scaling} illustrates how \demenba/ performance evolves as we vary (a) the amount of training data and (b) the maximum segment duration. In Fig.~\ref{fig:eff_of_sample_size}, we plot AUC as a function of total training hours per class, from 50 to 400 hours, for both 2-class and 3-class ADC.
Both \demenba/-small and \demenba/-medium show steadily increasing AUC up to 400 hours per class. Even at the largest data point we tested, the curves have not plateaued for \demenba/-medium, suggesting that additional hours would likely yield further gains.
The slope of AUC versus hours is steeper for 3-class ADC than 2-class ADC across both model sizes. This aligns with recent scaling-law observations in sequence modeling, where more complex tasks demand more data to resolve finer distinctions~\cite{kaplan2020scaling}. 
At each data level, \demenba/-medium outperforms \demenba/-small, and the gap widens as we increase training hours. Further, \demenba/-small exhibits diminishing returns after 200 hours per class, whereas \demenba/-medium continues improving beyond 400 hours/class. The difference suggests that \demenba/-medium better exploits additional examples rather than merely scaling its parameter count. Fig.~\ref{fig:eff_of_segment_duration} shows the relationship between maximum segment duration (from 30s to 360s) and AUC. 
Moving from 30s to 180s yields consistent AUC gains (e.g., 8-17\% absolute improvement in 2-class, 4-5\% in 3-class). This supports the hypothesis that dementia markers (hesitations, prosodic changes) often span multiple turns.
However, diminishing return occurs beyond 180s: going from 180s to 360s yields only a marginal 1-2 point AUC increase. Across both variants, 2-class AUC grows more sharply with segment duration than 3-class AUC.
\demenba/-medium's AUC increases by 17\% between 30s and 360s (2-class), whereas \demenba/-small improves by 8\%. This suggests that \demenba/-medium's deeper architecture better captures long-range dependencies.   

\begin{figure}[t]
    \centering
    \includegraphics[width=0.4\textwidth]{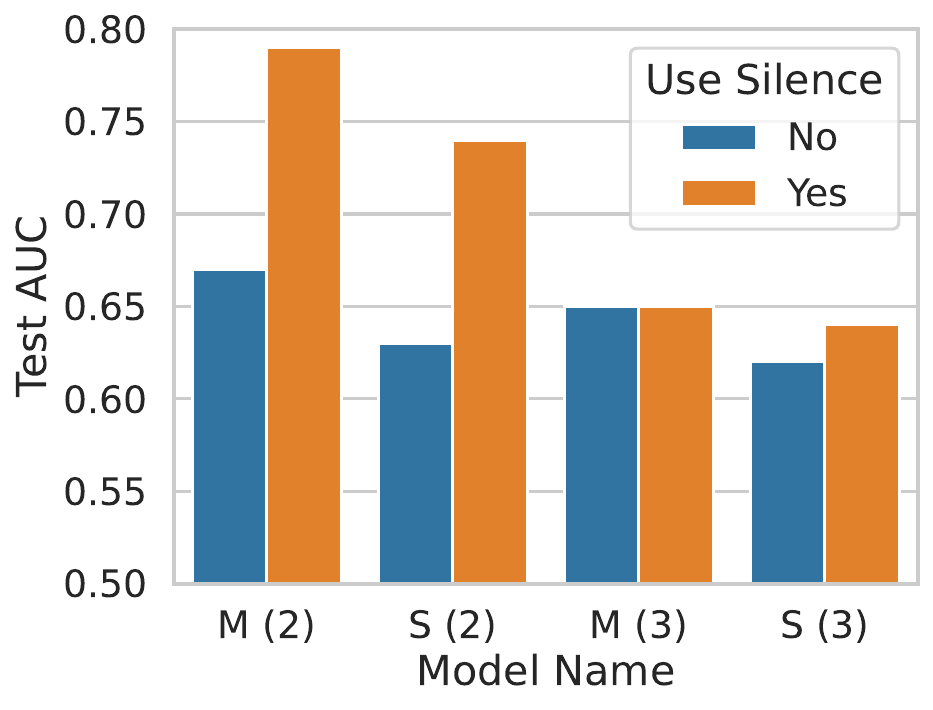}
    \caption{\textbf{Effect of including silences in audio on dementia classification performance for different model sizes and number of classes.} All models are trained with  100-hour/class subset and 360s segments. M ($k$) and S ($k$) stand for \demenba/-medium ($k$-class) and  \demenba/-small ($k$-class) respectively.}
    \label{fig:eff_of_silence}
\end{figure}
\begin{figure}[t]
    \centering
    \includegraphics[width=0.4\textwidth]{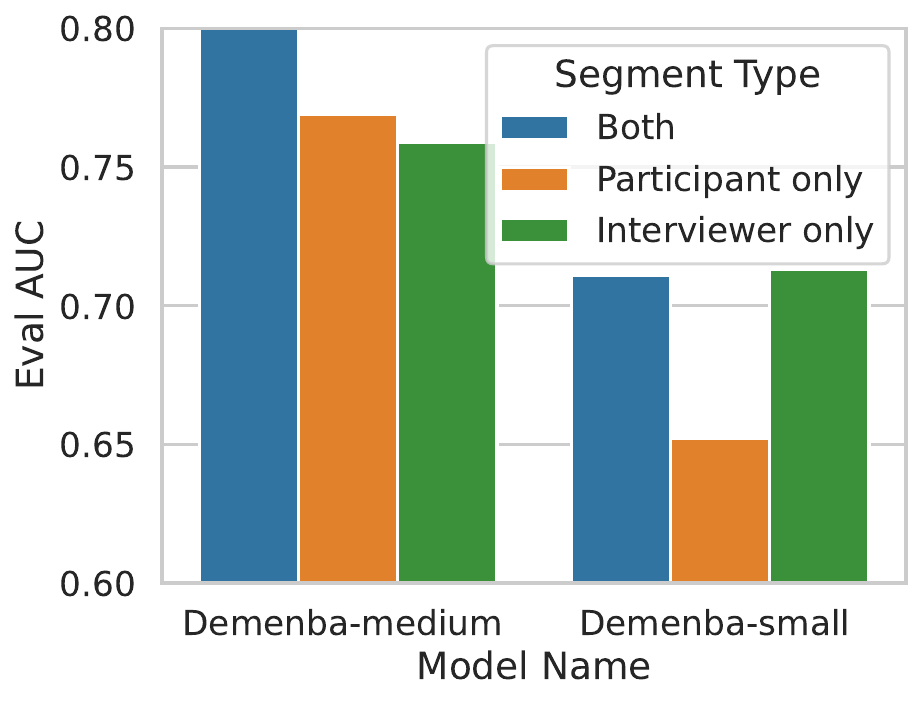}
    \caption{\textbf{Effect of including different speaker role segments for 2-class classification on the eval set.} All models are trained with 360-second segments.}
    \label{fig:eff_of_role_type}
\end{figure}
\subsection{Effect of segment types}
We next analyze how different segment types --- silence vs. non-silence, and interviewer vs. participant speech --- affect ADC performance.
Fig.~\ref{fig:eff_of_silence} compares AUC when training \demenba/ on only non-silence frames\liming{400-hour subset result?} versus combined speech+silence.
Including silence segments yields a 10-15\% AUC boost over speech-only models for the 2-class setting, probably because silence often signals hesitation and word-finding difficulty --- hallmarks of cognitive decline. 
In the 3-class setting, adding silence improves AUC by only 0-2\% over speech-only baselines. This suggests that while extreme silences help to discriminate ``normal'' from ``advanced dementia'', they are less helpful to distinguish between MCI and dementia due to similar hesitation patterns. Fig.~\ref{fig:eff_of_role_type} examines how model performance varies when training on participant segments only, interviewer segments only, or both. We use speaker boundaries derived from our SD system during training and gold boundaries to isolate each role's speech at test time. 
Training with both interviewer and participant speech yields the highest AUC of 0.81 for \demenba/-medium, about 4\% and 5\% higher than training with participant speech only and interviewer speech only respectively. On the other hand, \demenba/-small relies mostly on the interviewer segments, as even training on interviewer speech alone yields similar performance to training with both, and training on participant speech alone degrades the AUC by 6\%. In general, we found a significant amount of dementia-related information in the interviewer speech, consistent with prior works~\cite{al-hanai-etal-2018-role,perez-toro2021interviewer,dawalatabad-etal-2022-detecting} on unconscious interviewer bias toward impaired participants. 

\begin{figure}[t]
    \centering
    \includegraphics[width=0.42\textwidth]{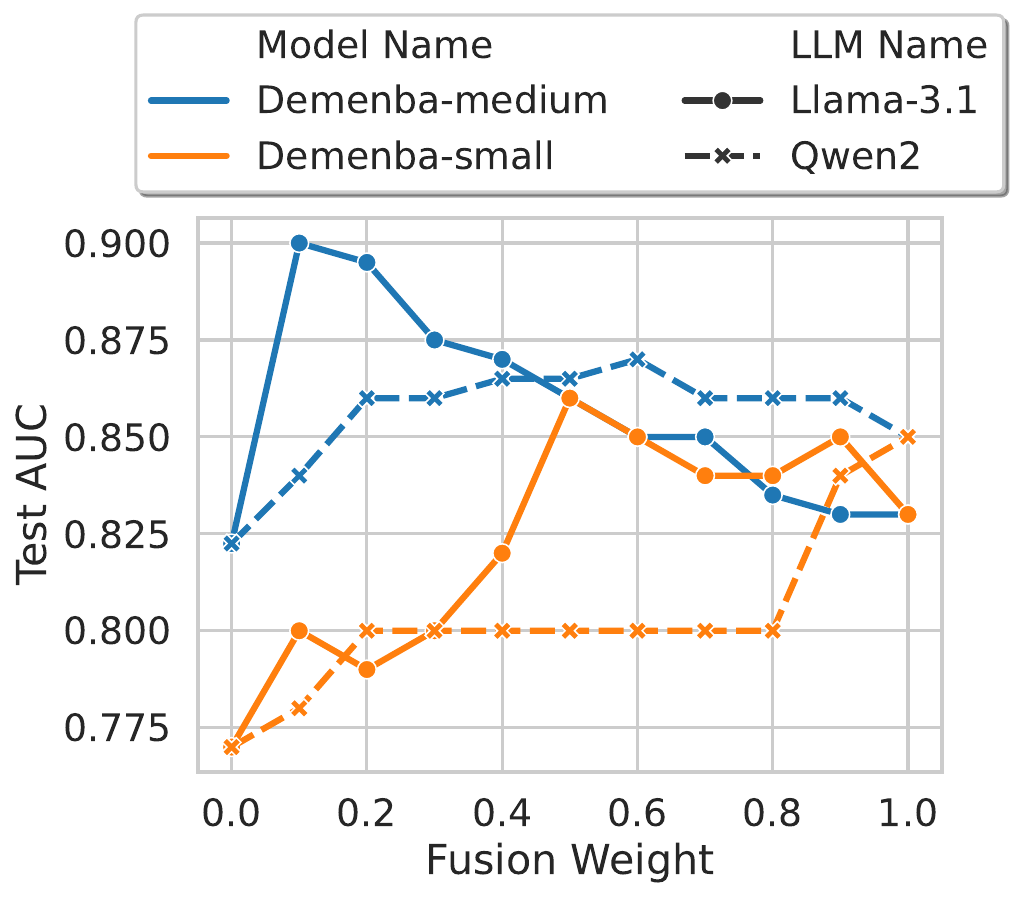}
    \caption{\textbf{Dementia classification performance vs. audio-text fusion weight for various audio and text models.} All models are trained with 360-second segments.}
    \label{fig:eff_of_fusion_weight}
\end{figure}
\subsection{Relative importance of linguistic vs. acoustic information}
 Fig.~\ref{fig:eff_of_fusion_weight} examines how AUC varies as we sweep the fusion weight between the audio-only \demenba/ and the text-only classifier, including LLaMA and \qwentwo/. We observe that the optimal fusion weight to be model-dependent. For \qwentwo/ (2-class), the peak AUC occurs at $\lambda\approx 0.5$ (50\% reliance on text cues), while for \llama/, the peak occurs at $\lambda\approx 0.1.$ For both variants of \demenba/, we observe a higher fusion weight or higher reliance on the text-based model for \qwentwo/ than for \llama/.\liming{TODO 3-class version? BERT?}

\section{Conclusion}

We study the use of SSMs for ADC from neuropsychological tests. Our method outperforms previous best methods, especially in the more challenging 3-class setting. Furthermore, we show that our approach is scalable to over 1,000 hours of speech data during training and long audio up to 6 minutes during both training and inference. Lastly, our analysis on the model sheds new lights on the role of acoustic features, speaker information, linguistic information and reasoning in the ADC task. In the future, we would like to further improve the performance and interpretability of our models by integrating it with multimodal LLMs. Other directions include studying more fine-grained classification of dementia subtypes and  generalization performance on other dementia datasets, including those with de-identified speech recordings.


\bibliographystyle{ieeetr}
\bibliography{reference,ref_ad,ref_ssm}

\begin{thebibliography}{10}

\bibitem{szekely2004nsaids}
C.~A. Szekely, J.~E. Thorne, P.~P. Zandi, M.~Ek, E.~Messias, J.~C. Breitner,
  and S.~N. Goodman, ``Nonsteroidal antiinflammatory drugs for the prevention
  of alzheimer’s disease: a systematic review,'' {\em Neuroepidemiology},
  vol.~23, no.~4, pp.~159--169, 2004.

\bibitem{chuang2016midlife}
Y.~F. Chuang, Y.~An, M.~Bilgel, D.~F. Wong, J.~C. Troncoso, R.~J. O’Brien,
  J.~C. Breitner, L.~Ferrucci, S.~M. Resnick, and M.~Thambisetty, ``Midlife
  adiposity predicts earlier onset of alzheimer’s dementia, neuropathology
  and presymptomatic cerebral amyloid accumulation,'' {\em Molecular
  Psychiatry}, vol.~21, pp.~910--915, 2016.

\bibitem{kurlowicz1999mmse}
L.~Kurlowicz and M.~Wallace, ``The mini-mental state examination (mmse),'' {\em
  Journal of Gerontological Nursing}, vol.~25, no.~5, pp.~8--9, 1999.

\bibitem{Luz2021_alzheimer_adresso}
S.~Luz, F.~Haider, S.~de~la Fuente, D.~Fromm, and B.~MacWhinney, ``Detecting
  cognitive decline using speech only: The {ADReSSo} challenge,'' in {\em
  Interspeech}, pp.~3780--3784, 2021.

\bibitem{alhanai2017spoken}
T.~Alhanai, R.~Au, and J.~Glass, ``Spoken language biomarkers for detecting
  cognitive impairment,'' in {\em 2017 IEEE Automatic Speech Recognition and
  Understanding Workshop (ASRU)}, pp.~409--416, 2017.

\bibitem{dawalatabad-etal-2022-detecting}
N.~Dawalatabad, Y.~Gong, S.~Khurana, R.~Au, and J.~Glass, ``Detecting dementia
  from long neuropsychological interviews,'' in {\em Findings of the
  Association for Computational Linguistics: EMNLP 2022}, (Abu Dhabi, United
  Arab Emirates), pp.~5270--5283, Association for Computational Linguistics,
  2022.

\bibitem{al-hanai-etal-2018-role}
T.~Al~Hanai, R.~Au, and J.~Glass, ``Role-specific language models for
  processing recorded neuropsychological exams,'' in {\em Proceedings of the
  2018 Conference of the North {A}merican Chapter of the Association for
  Computational Linguistics: Human Language Technologies, Volume 2 (Short
  Papers)} (M.~Walker, H.~Ji, and A.~Stent, eds.), (New Orleans, Louisiana),
  pp.~746--752, Association for Computational Linguistics, June 2018.

\bibitem{perez-toro2021interviewer}
P.~P{\'e}rez-Toro, S.~Bayerl, T.~Arias-Vergara, J.~V{\'a}squez-Correa,
  P.~Klumpp, M.~Schuster, E.~N{\"o}th, J.~Orozco-Arroyave, and K.~Riedhammer,
  ``Influence of the interviewer on the automatic assessment of alzheimer’s
  disease in the context of the adresso challenge,'' in {\em Proceedings of
  Interspeech}, pp.~3785--3789, 2021.

\bibitem{Vaswani2017}
Vaswani {\em et~al.}, ``Attention is all you need,'' in {\em NeurIPS},
  p.~6000–6010, 2017.

\bibitem{gu2021efficiently}
A.~Gu, K.~Goel, and C.~Ré, ``Efficiently modeling long sequences with
  structured state spaces,'' {\em arXiv preprint arXiv:2111.00396}, 2021.

\bibitem{Balagopalan2021-alzheimer-w2v2}
A.~Balagopalan and J.~Novikova, ``Comparing acoustic-based approaches for
  alzheimer’s disease detection,'' in {\em Interspeech}, pp.~3800--3804,
  2021.

\bibitem{Li2023-alzheimer-whisper}
J.~Li, K.~Song, J.~Li, B.~Zheng, D.~Li, X.~Wu, X.~Liu, and H.~Meng,
  ``Leveraging pretrained representations with task-related keywords for
  alzheimer's disease detection,'' in {\em ArXiv}, 2023.

\bibitem{bhati2024dass}
S.~Bhati, Y.~Gong, L.~Karlinsky, H.~Kuehne, R.~Feris, and J.~Glass, ``Dass:
  Distilled audio state space models are stronger and more duration-scalable
  learners,'' in {\em SLT}, 2024.

\bibitem{gu2022efficiently}
A.~Gu, K.~Goel, and C.~R{\'e}, ``Efficiently modeling long sequences with
  structured state spaces,'' {\em ICLR}, 2022.

\bibitem{gu2023mamba}
A.~Gu and T.~Dao, ``Mamba: Linear-time sequence modeling with selective state
  spaces,'' {\em arXiv preprint arXiv:2312.00752}, 2023.

\bibitem{Luz2020_alzheimer_adress}
S.~Luz, F.~Haider, S.~de~la Fuente, D.~Fromm, and B.~MacWhinney,
  ``{Alzheimer's} dementia recognition through spontaneous speech: The {ADReSS
  Challenge},'' in {\em Interspeech}, (Shanghai, China), 2020.

\bibitem{Rohanian2021-alzheimer-speech}
M.~Rohanian, J.~Hough, and M.~Purver, ``Alzheimer’s dementia recognition
  using acoustic, lexical, disfluency and speech pause features robust to noisy
  inputs,'' in {\em Interspeech}, pp.~3820--3824, 2021.

\bibitem{xue2021dementia}
C.~Xue, C.~Karjadi, I.~C. Paschalidis, R.~Au, and V.~B. Kolachalama,
  ``Detection of dementia on voice recordings using deep learning: a framingham
  heart study,'' {\em Alzheimer’s Research \& Therapy}, vol.~13, p.~146, Aug.
  2021.

\bibitem{Baevski2020-wav2vec2}
A.~Baevski, H.~Zhou, A.~Mohamed, and M.~Auli, ``wav2vec 2.0: A framework for
  self-supervised learning of speech representations,'' in {\em NeurIPS}, 2020.

\bibitem{Radford2023-whisper}
A.~Radford, J.~W. Kim, T.~Xu, G.~Brockman, C.~McLeavey, and I.~Sutskever,
  ``Robust speech recognition via large-scale weak supervision,'' in {\em
  ICML}, 2023.

\bibitem{Haulcy2021-alzheimer-speech}
R.~Haulcy and J.~Glass, ``Classifying alzheimer's disease using audio and
  text-based representations of speech,'' {\em Frontiers in Psychology},
  vol.~11, p.~624137, 2021.

\bibitem{liu2024vmamba}
Y.~Liu, Y.~Tian, Y.~Zhao, H.~Yu, L.~Xie, Y.~Wang, Q.~Ye, and Y.~Liu, ``Vmamba:
  Visual state space model,'' {\em arXiv preprint arXiv:2401.10166}, 2024.

\bibitem{zhu2024vision}
L.~Zhu, B.~Liao, Q.~Zhang, X.~Wang, W.~Liu, and X.~Wang, ``Vision mamba:
  Efficient visual representation learning with bidirectional state space
  model,'' {\em arXiv preprint arXiv:2401.09417}, 2024.

\bibitem{erol2024audio}
M.~H. Erol, A.~Senocak, J.~Feng, and J.~S. Chung, ``Audio mamba: Bidirectional
  state space model for audio representation learning,'' {\em IEEE Signal
  Processing Letters}, vol.~31, pp.~2975--2979, 2024.

\bibitem{lin2024audio}
J.~Lin and H.~Hu, ``Audio mamba: Pretrained audio state space model for audio
  tagging,'' {\em arXiv preprint arXiv:2405.13636}, 2024.

\bibitem{shams2024ssamba}
S.~Shams, S.~S. Dindar, X.~Jiang, and N.~Mesgarani, ``Ssamba: Self-supervised
  audio representation learning with mamba state space model,'' {\em arXiv
  preprint arXiv:2405.11831}, 2024.

\bibitem{zhang2025mamba}
X.~Zhang, Q.~Zhang, H.~Liu, T.~Xiao, X.~Qian, B.~Ahmed, E.~Ambikairajah, H.~Li,
  and J.~Epps, ``Mamba in speech: Towards an alternative to self-attention,''
  {\em IEEE Transactions on Audio, Speech and Language Processing}, 2025.

\bibitem{jiang2025speech}
X.~Jiang, Y.~A. Li, A.~N. Florea, C.~Han, and N.~Mesgarani, ``Speech slytherin:
  Examining the performance and efficiency of mamba for speech separation,
  recognition, and synthesis,'' in {\em ICASSP 2025-2025 IEEE International
  Conference on Acoustics, Speech and Signal Processing (ICASSP)}, pp.~1--5,
  IEEE, 2025.

\bibitem{devlin2019bert}
J.~Devlin, M.-W. Chang, K.~Lee, and K.~Toutanova, ``{BERT}: Pre-training of
  deep bidirectional transformers for language understanding,'' in {\em
  Proceedings of the 2019 Conference of the North American Chapter of the
  Association for Computational Linguistics: Human Language Technologies
  (NAACL-HLT)}, pp.~4171--4186, Association for Computational Linguistics,
  2019.

\bibitem{Plaquet23}
A.~Plaquet and H.~Bredin, ``{Powerset multi-class cross entropy loss for neural
  speaker diarization},'' in {\em Proc. INTERSPEECH 2023}, 2023.

\bibitem{Bredin23}
H.~Bredin, ``{pyannote.audio 2.1 speaker diarization pipeline: principle,
  benchmark, and recipe},'' in {\em Proc. INTERSPEECH 2023}, 2023.

\bibitem{deng2009imagenet}
J.~Deng, W.~Dong, R.~Socher, L.-J. Li, K.~Li, and L.~Fei-Fei, ``Imagenet: A
  large-scale hierarchical image database,'' in {\em 2009 IEEE Conference on
  Computer Vision and Pattern Recognition (CVPR)}, pp.~248--255, IEEE, 2009.

\bibitem{gemmeke2017audio}
J.~F. Gemmeke, D.~P.~W. Ellis, D.~Freedman, A.~Jansen, W.~Lawrence, R.~C.
  Moore, M.~Plakal, and M.~Ritter, ``Audio set: An ontology and human-labeled
  dataset for audio events,'' in {\em Proc. IEEE International Conference on
  Acoustics, Speech and Signal Processing (ICASSP)}, (New Orleans, LA, USA),
  IEEE, 2017.

\bibitem{tan2019efficientnet}
M.~Tan and Q.~V. Le, ``Efficientnet: Rethinking model scaling for convolutional
  neural networks,'' in {\em ICML}, vol.~97, pp.~6105--6114, PMLR, 2019.

\bibitem{touvron2023llama}
H.~Touvron, T.~Lavril, G.~Izacard, X.~Martinet, M.-A. Lanchaux, T.~Lacroix,
  B.~Rozi{\`e}re, N.~Goyal, E.~Hambro, F.~Azhar, A.~Rodriguez, A.~Joulin,
  E.~Grave, and G.~Lample, ``{LLaMA}: Open and efficient foundation language
  models,'' {\em arXiv preprint arXiv:2302.13971}, 2023.

\bibitem{yang2024_qwen2}
A.~Yang {\em et~al.}, ``Qwen2 technical report,'' tech. rep., Qwen Team,
  Alibaba Group, 2024.
\newblock Technical Report.

\bibitem{phi4mini2024technical}
Microsoft, ``Phi-4-mini technical report: Compact yet powerful multimodal
  language models via mixture-of-loras,'' {\em Technical Report}, 2024.

\bibitem{hu2021lora}
E.~J. Hu, Y.~Shen, P.~Wallis, Z.~Allen-Zhu, Y.~Li, L.~Wang, and W.~Chen,
  ``Lora: Low-rank adaptation of large language models,'' 2021.

\bibitem{kingma2015adam}
D.~P. Kingma and J.~Ba, ``Adam: A method for stochastic optimization,'' in {\em
  Proceedings of the 3rd International Conference on Learning Representations
  (ICLR)}, (San Diego, CA), 2015.

\bibitem{kaplan2020scaling}
J.~Kaplan, S.~McCandlish, T.~Henighan, T.~B. Brown, B.~Chess, R.~Child,
  S.~Gray, A.~Radford, J.~Wu, and D.~Amodei, ``Scaling laws for neural language
  models,'' {\em arXiv preprint arXiv:2001.08361}, 2020.

\end{thebibliography}

\end{document}